\documentclass{article}

\usepackage[preprint]{neurips_2024}

\usepackage[dvipsnames]{xcolor}
\usepackage{tikz}
\usetikzlibrary{backgrounds}
\usetikzlibrary{arrows,shapes}
\usetikzlibrary{tikzmark}
\usetikzlibrary{calc}
\usepackage{blindtext}
\usepackage{xspace}
\usepackage{array}
\usepackage{ragged2e}
\newcolumntype{P}[1]{>{\RaggedRight\hspace{0pt}}p{#1}}
\newcolumntype{X}[1]{>{\RaggedRight\hspace*{0pt}}p{#1}}

\usepackage{tcolorbox}

\usepackage{tikz}
\usetikzlibrary{arrows,shapes,positioning,shadows,trees,mindmap}
\usepackage[edges]{forest}
\usetikzlibrary{arrows.meta}
\colorlet{linecol}{black!75}
\usepackage{xkcdcolors} 

\usepackage{tikz}
\usetikzlibrary{backgrounds}
\usetikzlibrary{arrows,shapes}
\usetikzlibrary{tikzmark}
\usetikzlibrary{calc}

\colorlet{mhpurple}{Plum!80}

\usepackage[utf8]{inputenc} 
\usepackage[T1]{fontenc}    
\usepackage{xr-hyper}
\usepackage[pagebackref=true,breaklinks=true,colorlinks=true,linkcolor=red,citecolor=cyan,filecolor=magenta,urlcolor=cyan,bookmarks=false]{hyperref}

\usepackage{multirow}
\usepackage{graphicx}
\usepackage{amsmath}
\usepackage{amssymb}
\usepackage{mathtools, nccmath}
\usepackage{amsthm}
\usepackage{pifont}

\usepackage{algorithm, algorithmic}
\usepackage{wrapfig, lipsum, booktabs}
\usepackage{colortbl}
\usepackage{comment}
\usepackage{cleveref}
\usepackage{marvosym}      
\usepackage{subcaption}
 \usepackage{enumitem}

\title{Is a 3D-Tokenized LLM the Key to Reliable Autonomous Driving?}

\author{
  Yifan Bai\textsuperscript{1}\thanks{ Equal contributions. $^\dagger$This work was done during internship at MEGVII Technology. $^\ddagger$Corresponding authors. The work was supported by National Science and Technology Major Project of China (2023ZD0121300) and CAAI-MindSpore Open Found, the FDCT grants 0102/2023/RIA2, 0154/2022/A3.}\textsuperscript{\ \ $\dagger$}\quad Dongming Wu\textsuperscript{2$*$$\dagger$}\quad Yingfei Liu\textsuperscript{3}\quad Fan Jia\textsuperscript{3}\quad Weixin Mao\textsuperscript{3}\quad Ziheng Zhang\textsuperscript{3}\quad \\ \textbf{Yucheng Zhao\textsuperscript{3}\quad Jianbing Shen\textsuperscript{4$\ddagger$}\quad Xing Wei\textsuperscript{1$\ddagger$}\quad Tiancai Wang\textsuperscript{3$\ddagger$}\quad Xiangyu Zhang}\textsuperscript{3} \\
  \textsuperscript{1} Xi'an Jiaotong University, 
  \textsuperscript{2} Beijing Institute of Technology, \\ 
  \textsuperscript{3} MEGVII Technology,
  \textsuperscript{4} SKL-IOTSC, University of Macau \\
\texttt{yfbai@stu.xjtu.edu.cn, \{wudongming97, shenjianbingcg\}@gmail.com}, \\
\texttt{weixing@mail.xjtu.edu.cn, \{liuyingfei,wangtiancai\}@megvii.com} 
}

\begin{document}

\maketitle

\begin{abstract}
Rapid advancements in Autonomous Driving (AD) tasks turned a significant shift toward end-to-end fashion, particularly in the utilization of vision-language models (VLMs) that integrate robust logical reasoning and cognitive abilities to enable comprehensive end-to-end planning.
However, these VLM-based approaches tend to integrate 2D vision tokenizers and a large language model (LLM) for ego-car planning, which lack 3D geometric priors as a cornerstone of reliable planning. 
Naturally, this observation raises a critical concern: \textbf{Can a 2D-tokenized LLM accurately perceive the 3D environment?} Our evaluation of current VLM-based methods across 3D object detection, vectorized map construction, and environmental caption suggests that the answer is, unfortunately, \textbf{NO}. 
In other words, 2D-tokenized LLM fails to provide reliable autonomous driving. 
In response, we introduce DETR-style 3D perceptrons as 3D tokenizers, which connect LLM with a one-layer linear projector. This simple yet elegant strategy, termed Atlas, harnesses the inherent priors of the 3D physical world, enabling it to simultaneously process high-resolution multi-view images and employ spatiotemporal modeling.
Despite its simplicity, Atlas demonstrates superior performance in both 3D detection and ego planning tasks on nuScenes dataset, proving that 3D-tokenized LLM is the key to reliable autonomous driving. The code and datasets will be released.

\end{abstract}
\section{Introduction}
\label{sec:intro}

\begin{figure}[t]
    \centering
    \includegraphics[width=\linewidth]{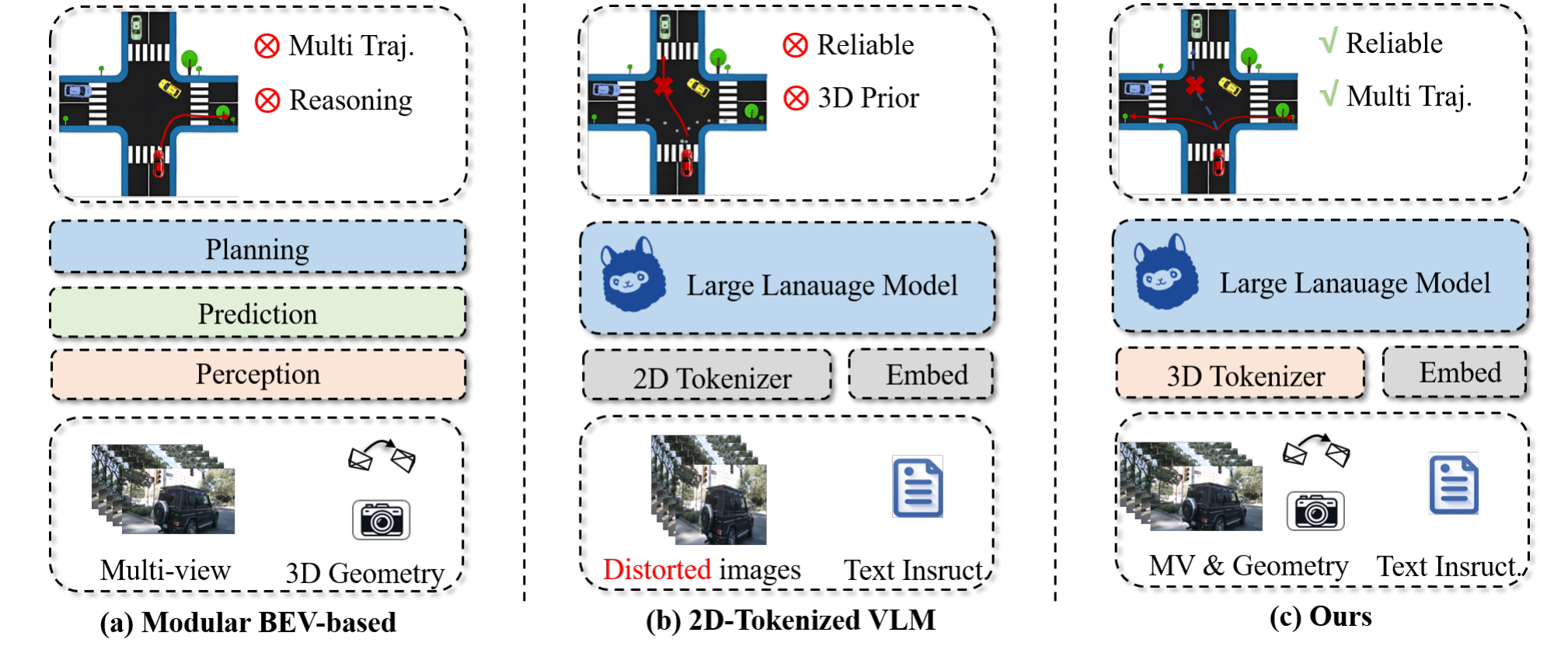}
    \vspace{-20pt}
    \caption{\textbf{Comparision among end-to-end methods}. (a) Modular BEV-based methods have three sequential modules for perception, prediction, and planning, but they cannot provide multiple potential trajectories and environment reasoning. (b) 2D-tokenized VLM projects 2D distorted images into tokens, which lack 3D prior for reliable autonomous driving. (c) Our 3D-tokenized LLM-based methods utilize 3D perceptions as 3D tokenizers, which provide potential trajectories and rich 3D priors for reliable driving.}
    \label{fig:introduction}
\end{figure}


Autonomous Driving (AD) is a sophisticated system that integrates perception, reasoning, and planning~\cite{pomerleau1988alvinn,janai2020computer,chen2023end}. 
Perception serves as the initial stage, capturing details of the surrounding environment. This information then feeds into the reasoning component, facilitating a deeper understanding, and ultimately guiding informed decision-making through the planning process. 
Recently, the incorporation of perception, reasoning, and planning to construct end-to-end models has become prevalent.
It can be broadly categorized into two distinct methodologies: modular bird's-eye view (BEV) based approaches and large vision-language model (VLM) based methods.

The modular BEV-based approaches are meticulously engineered, comprising custom-tailored modules, including 3D perception, trajectory prediction, and ego-car planning~\cite{liang2020pnpnet,casas2021mp3,chen2022learning,zhang2022beverse,hu2022st,gu2023vip3d,hu2023planning}, as shown in Figure~\ref{fig:introduction}(a).
While BEV representation enhances environmental perception, these methods may encounter difficulty stemming from their limited reasoning abilities.
Specifically, these models tend to mimic established expert trajectories and struggle to predict multiple potential motion trajectories when confronted with novel scenarios. 
To tackle this challenge, VLM-based methods mark a significant turning point.
They usually employ a 2D vision tokenizer ({e.g.}, ViT-CLIP~\cite{radford2021learning}) with a Large Language Model (LLM) to interpret distorted images and produce navigational commands~\cite{xu2023drivegpt4,tian2024drivevlm,wang2023drivemlm,shao2023lmdrive,jia2023adriver,sima2023drivelm}. Benefiting from the robust logical reasoning and cognitive abilities of the VLM agent, the model can generate rational decisions and dialogues.

Despite the success of VLM-based algorithms, the perceptual capabilities within this paradigm is barely studied. While we argue that the perception sub-task may not be essential for end-to-end driving, the capacity to perceive the environment remains a cornerstone of reliable planning. Since VLM-based methods rely on 2D vision tokenizers for environmental perception without incorporating 3D geometric priors, an intuitive question arises:
\textbf{Can a 2D-tokenized LLM accurately perceive the 3D environment?} To answer this question, we specially design experiments to evaluate the perception performance of prevalent VLM-based systems in three tasks: 3D object detection, 3D lane detection, and environmental captioning.  
Our findings reveal that despite extensive pre-training and expansive parameters, mainstream VLM solutions typically lag in precision when compared to specialized models designed for these tasks. This glaring gap highlights the limitations of 2D tokenizers in perceiving 3D environments.

To address this issue, we wonder if 3D vision tokenizers hold the key to Pandora. We discover that the existing DETR-style BEV framework can naturally serve as a 3D visual compression tokenizer. 
Therefore, we opt for the advanced StreamPETR~\cite{wang2023exploring} and TopoMLP~\cite{wu2023topomlp} as our 3D visual tokenizers, forgoing the traditional use of ViT-CLIP~\cite{radford2021learning}.
This strategy brings three advantages: \textbf{1)} The innate priors of the 3D physical world are naturally encoded within visual tokens by introducing the position encodings. \textbf{2)} It is capable of handling high-resolution images with any aspect ratio without the risk of distorting the images. \textbf{3)} Video frames can be processed in a streaming manner, benefiting from DETR-style query propagation. Through evaluation of the nuScenes dataset, we demonstrate that our 3D-tokenized LLM approach achieves performance on par with specialized algorithms in tasks such as 3D object detection and lane detection.

Beyond that, we need to answer another question: \textbf{Is a 3D-tokenized LLM the key to reliable autonomous driving?} Following BEV-Planner~\cite{li2023ego}, we extend our exploration to the open-loop planning on the nuScenes dataset.  By leveraging the 3D tokenizers for enhanced perception capabilities, our model not only comprehends the environment around the vehicle but also utilizes the LLM to formulate driving recommendations and plan the ego-car trajectory in an end-to-end manner. Remarkably, this approach eschews hand-crafted designs and achieves state-of-the-art performance on the nuScenes planning task. 

In summary, our work highlights the importance of proper vision tokenizers in VLM-based AD and introduces the 3D-tokenized LLM as a solution. We showcase its superiority in adeptly addressing challenges across multiple tasks such as 3D perception, vectorized map construction, environmental caption, and planning within autonomous driving systems. Our model demonstrates superior performance in both benchmark evaluations and practical downstream applications, proving its reliability and versatility. Furthermore, our framework paves the way for pioneering end-to-end LLM-driven solutions in autonomous driving, potentially transforming how these systems are developed.

\section{Can a 2D-Tokenized LLM Accurately Perceive 3D Environment?}
Current VLM-based methods~\cite{xu2023drivegpt4,tian2024drivevlm,wang2023drivemlm,shao2023lmdrive,jia2023adriver} in AD tend to employ 2D vision tokenizers.
They operate without incorporating geometric 3D priors, raising concerns about their capability to accurately perceive and describe 3D environments, which is crucial for reliable planning. 
In this section, we provide insightful analysis and reveal the limitations of relying solely on 2D tokenizers for understanding 3D driving scenes, including 3D perception and visual captioning.

\subsection{2D-Tokenized LLM for Perception}

To investigate the 3D understanding capability of current VLM-based approaches, we first conduct experiments on traditional perception tasks: 3D object detection and 3D lane detection.
In this part, we introduce datasets, models, and metrics.

\begin{figure}[t]
    \centering
    \includegraphics[width=0.925\linewidth]{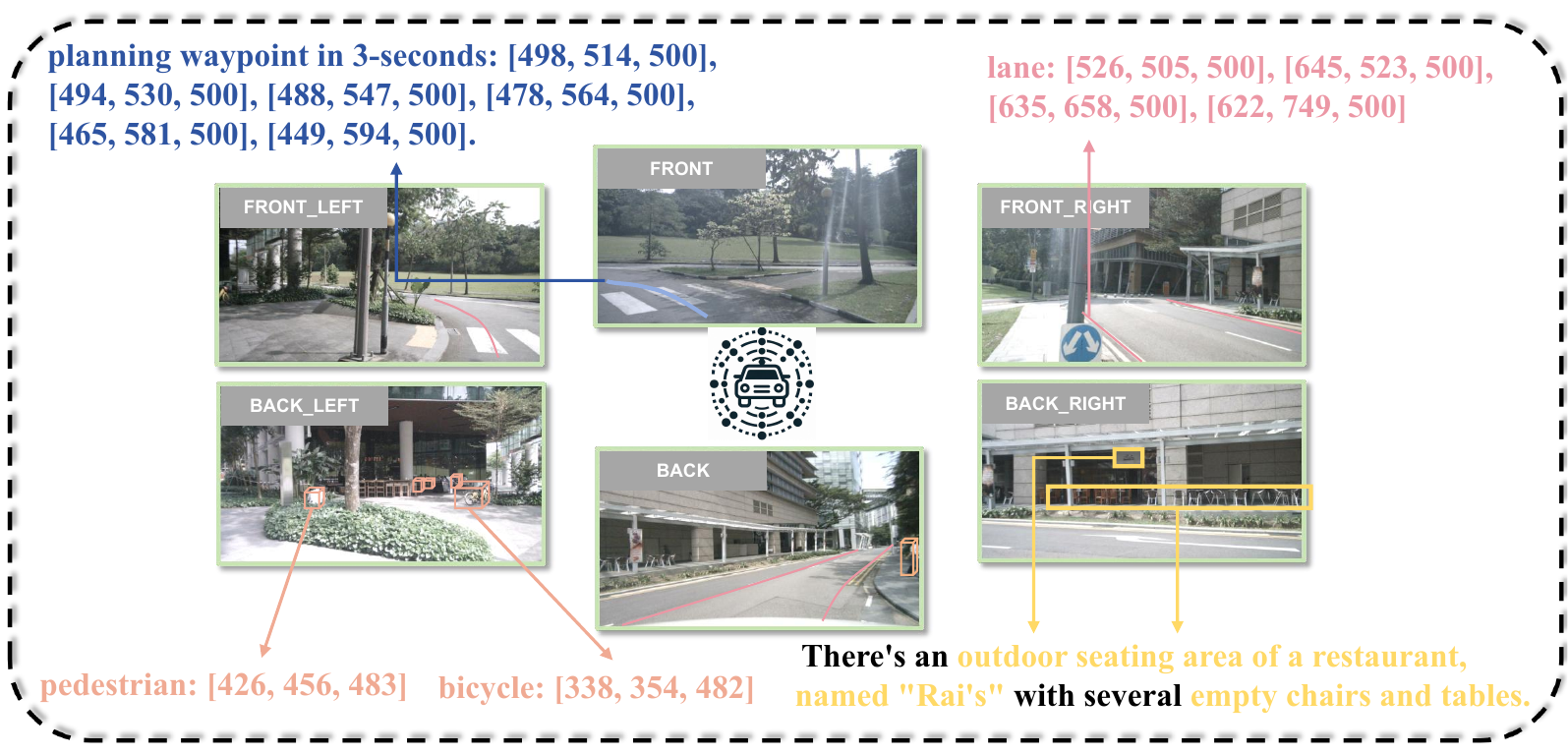}
    \vspace{-4pt}
    \caption{\textbf{Brief answer format of datasets.} It transforms several tasks, such as 3D object detection, map perception, environment caption, and ego-car planning, into a uniform text format. We discretize the bird's-eye view (BEV) space, spanning from -50 meters to +50 meters, into 1,000 bins.}
    \label{fig:datasets}
\end{figure}

\textbf{Datasets.} 
We design datasets tailored for VLM methods built upon popular multi-view benchmark nuScenes~\cite{caesar2020nuscenes}, as shown in Figure~\ref{fig:datasets}.
For the 3D detection task, we construct question-and-answer (QA) pairs that focus on pinpointing the locations of objects surrounding the ego vehicle. 
Each question prompts the model to extract spatial information about the target objects from six views. The corresponding answers require the model to identify both the category and the 3D coordinates of objects. 
Similarly, the dataset for 3D lane detection also comprises QA pairs, whose answers are lane points borrowed from OpenLane-V2 subset-B~\cite{wang2024openlane}. Here, each road is depicted using four consecutive points describing the road centerline. 
More details can be found in the supplementary.

\textbf{Models.} 
All 2D-tokenized LLMs in our study adhere to a uniform architecture, which consists of three main components: 2D tokenizer, projector, and large language model. 
The 2D tokenizer follows ViT-CLIP~\cite{radford2021learning} to extract visual features from multiple perspectives of images.
For the projection module, we incorporate a single convolutional layer to bridge the 2D tokenizer and LLM. 
Besides, we utilize diverse pre-trained LLMs, such as LLaMA~\cite{touvron2023llama}, LLaVA~\cite{liu2024visual}, Vicuna~\cite{chiang2023vicuna}, which are comprehensive processing of complex visual information to generate the perception of the environment, to prove consistency and fairness in our exploration.
Additionally, another available VLM-based model pre-trained on 2D object detection Merlin~\cite{yu2023merlin} is also evaluated.

\textbf{Metrics.} 
In this study, we employ the F1 score as the main evaluation metric.
The choice of the F1 score is motivated by two primary considerations:
First, VLMs cannot deliver the necessary predictive confidence for metrics such as mean Average Precision (mAP). 
Second, traditional perceptual metrics commonly encourage numerous redundant predictions, which can clutter the model output. In contrast, VLMs are designed to generate more targeted and focused predictions, making the F1 score a better fit for assessing these models.
In this work, for 3D detection,  we choose threshold distances of 0.5, 1.0, 2.0, and 4.0 meters to define positive predictions, similar to the discrimination levels used in detection mAP calculations. 
As for 3D lane detection, we follow OpenLane-V2 evaluation protocol~\cite{wang2024openlane} to compute the F1 score.

\begin{table*}[t]
\centering
    \caption{Comparisons with task-specific and VLM-based methods for 3D object detection tasks using our proposed dataset. The bold \textbf{numbers} represent the highest accuracy achieved in each category. The P$_{k}$, R$_{k}$, and F1$_{k}$ represent the Precision, Recall, and respective F1 score ultimate $k$ as threshold distances to define positive prediction. The Spe. represents task-specialist model.}
\resizebox{\linewidth}{!}{
\begin{tabular}{cc|c|ccc|ccc|ccc|ccc}
    \toprule
     & \multicolumn{1}{c|}{Method} & Tokenizers &
 \textbf{P$_{0.5}$} & \textbf{R$_{0.5}$} & \textbf{F1$_{0.5}$} & \textbf{P$_{1.0}$} & \textbf{R$_{1.0}$} & \textbf{F1$_{1.0}$} & \textbf{P$_{2.0}$} & \textbf{R$_{2.0}$} & \textbf{F1$_{2.0}$} & \textbf{P$_{4.0}$} & \textbf{R$_{4.0}$} & \textbf{F1$_{4.0}$} \\
    \midrule
    \multirow{2}{*}{\rotatebox[origin=c]{90}{Spe.}} &
    PETR~\cite{liu2022petr} & - & 12.4 & 21.5 & 15.8 & 20.0 & 30.5 & 24.1 & 27.5 & 37.7 & 31.8 & 33.8 & 42.6 & 37.7 \\
& StreamPETR~\cite{wang2023exploring} & - & \textbf{22.7} & \textbf{41.3} & \textbf{29.3} & \textbf{31.6} & \textbf{49.5} & \textbf{38.6} & \textbf{38.1} & \textbf{54.2} & \textbf{44.7} & \textbf{42.5} & \textbf{56.9} & \textbf{48.7} \\
    \midrule
    \multirow{5}{*}{\rotatebox[origin=c]{90}{VLM}} & 
    LLaMA~\cite{touvron2023llama} & 2D & 0.3 & 1.1 & 0.4 & 0.6 & 2.6 & 1.0 & 1.5 & 5.8 & 2.4 & 3.5 & 12.8 & 5.5 \\
    & LLaVA~\cite{liu2024visual}& 2D & 2.0 & 20.3 & 3.0 & 3.6 & 35.7 & 6.5 & 6.5& 50.3 & 11.6 & 10.9 & 62.8 & 18.9 \\
    & Vicuna~\cite{chiang2023vicuna} & 2D & 2.0 & 20.1 & 2.5 & 2.9 & 35.6 & 5.4 & 5.9 & 51.1 & 10.1 & 9.4 & 63.8 & 16.4 \\
    & Merlin~\cite{yu2023merlin} & 2D & 3.0 & 22.5 & 5.3 & 4.1& 36.1 & 7.4 & 6.6 & 52.6 & 11.7 & 12.1 & 64.3 & 20.4 \\
    & \cellcolor{gray!15}\textbf{Atlas}(Ours) &\cellcolor{gray!15}\textbf{3D} & \cellcolor{gray!15}\textbf{15.0} & \cellcolor{gray!15}\textbf{61.2} & \cellcolor{gray!15}\textbf{24.1} & \cellcolor{gray!15}\textbf{27.2} & \cellcolor{gray!15}\textbf{74.0} & \cellcolor{gray!15}\textbf{39.8} & \cellcolor{gray!15}\textbf{36.2} & \cellcolor{gray!15}\textbf{79.2} & \cellcolor{gray!15}\textbf{49.7} & \cellcolor{gray!15}\textbf{41.2} & \cellcolor{gray!15}\textbf{81.2} & \cellcolor{gray!15}\textbf{54.6} \\
    \bottomrule
    \end{tabular}
    }
    \vspace{-5pt}
    \label{tbl:detection}
\end{table*}

\textbf{3D Object Detection.} 
In this study, we conduct extensive experiments to evaluate the performance of VLMs on 3D detection, as listed in Table \ref{tbl:detection}. 
As a comparison, Table \ref{tbl:detection} also includes task-specific models such as PETR~\cite{liu2022petr} and StreamPETR~\cite{wang2023exploring}. Among these, the state-of-the-art detector StreamPETR achieves an F1$_{4.0}$ score of 48.7.
Despite the rich contextual knowledge and extensive parameters, 2D-tokenized LLM methods exhibit a considerable performance drop in both precision and recall, leading to surprisingly low F1 scores.
These methods struggle with detecting objects in the vicinity of the ego vehicle, highlighting a considerable disparity in 3D object detection capabilities between VLM-based methods and dedicated task-specific approaches.

\setlength\intextsep{0pt}
\begin{wraptable}{h}{0.4\linewidth}
\footnotesize
\vspace{0mm}
\centering
\caption{3D lane detection.}  
\vspace{-2.5mm}
	   \setlength{\abovecaptionskip}{-0.1cm}
	   \setlength{\belowcaptionskip}{0.1cm}
         \centering
  	\resizebox{0.4\textwidth}{!}{
		\setlength\tabcolsep{6pt}
		\renewcommand\arraystretch{1}
		\begin{tabular}{c|c|ccc}
    \toprule		
    Method & Tokenizers & \textbf{P}  & \textbf{R} & \textbf{F1} \\ 
    \midrule
    TopoMLP~\cite{wu2023topomlp} & - & 50.6 & 55.7 & 53.0 \\
    \midrule
    LLaVA~\cite{liu2024visual} & 2D & 10.4& 9.8 & 10.0\\
    Vicuna~\cite{chiang2023vicuna} & 2D & 11.7 & 10.3 & 10.9 \\
    Merlin~\cite{yu2023merlin} & 2D & 22.1 & 22.4 & 22.2 \\
    \cellcolor{gray!15}\textbf{Atlas}(ours) &  \cellcolor{gray!15}\textbf{3D} & \cellcolor{gray!15}45.7 & \cellcolor{gray!15}39.1 & \cellcolor{gray!15}42.2 \\
    \bottomrule
    \end{tabular}}
\label{tbl:map}
\end{wraptable}

\textbf{3D Lane Detection.} 
Vectorized maps provide a driving route for ego car,  serving as a crucial perception task for autonomous driving.
We present experiments of a state-of-the-art task-specific model TopoMLP~\cite{wu2023topomlp} and several aforementioned 2D-tokenized LLM methods on lane detection. 
The main results are shown in Table~\ref{tbl:map}.
Similarly, the performance of 2D-tokenized LLM methods is far away from the task-specific model, struggling to deal with 3D lane detection.


\subsection{2D-Tokenized LLM for Captioning}

\begin{figure}[t]
    \centering
    \includegraphics[width=\linewidth]{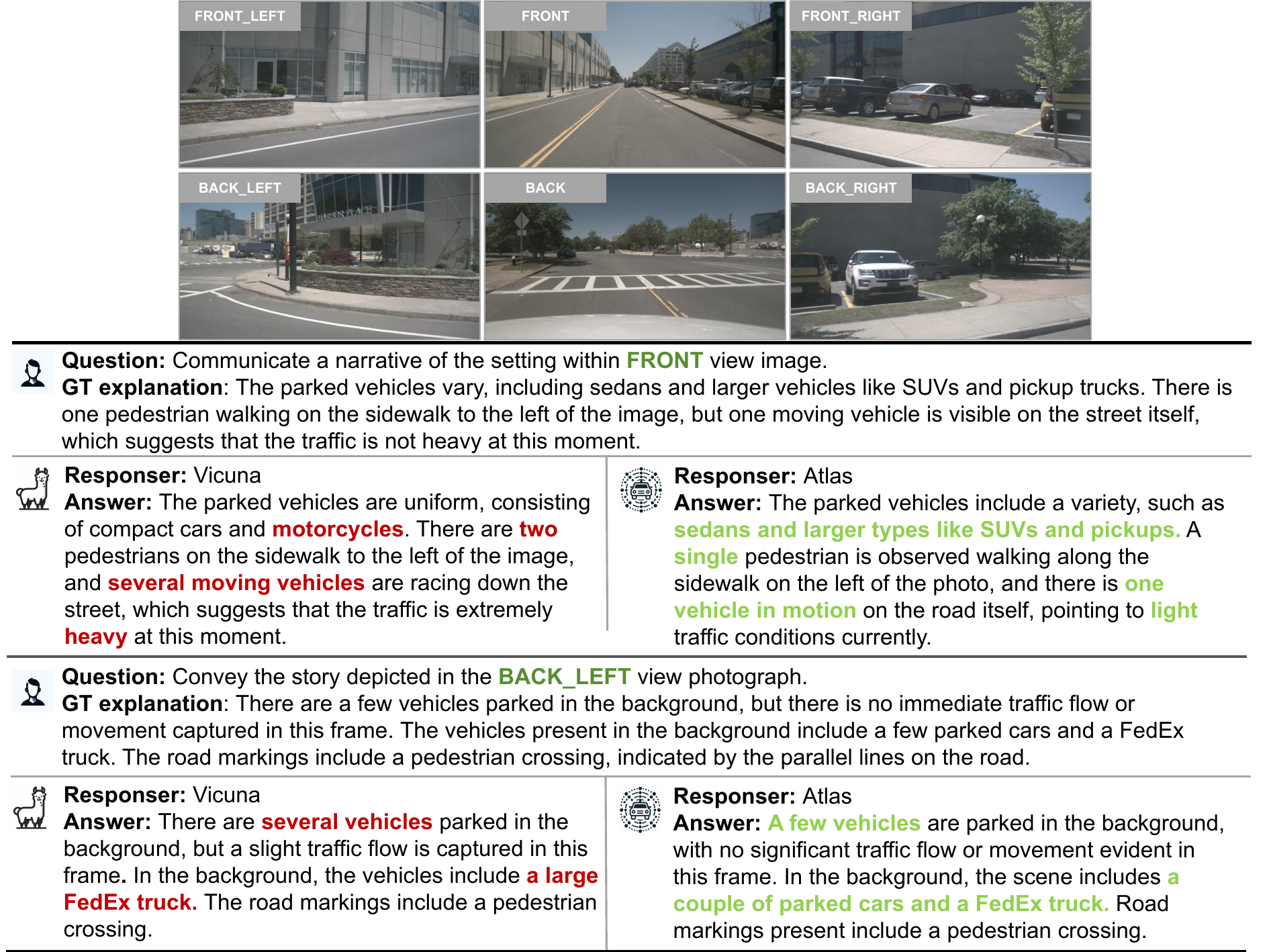}
    \vspace{-16pt}
    \caption{\textbf{Comparsion between 2D-tokenized and our 3D-tokenized VLMs on driving caption.} The 2D-tokenized VLM sometimes generates "hallucinated" descriptions, while our 3D-tokenized VLM is able to produce accurate and comprehensive captions for driving environment.}
    \label{fig:caption}
\end{figure}

In addition to basic environmental perception tasks, LLMs can be adapted to perform more complex tasks like extracting and interpreting key features from visual for captioning environments. This capability extends the utility of LLMs in practical applications, and leverages world knowledge and reasoning ability, particularly in scenarios requiring detailed environmental understanding.

To explore whether a 2D-tokenized LLM could serve as an effective perceptron, we develop a specialized version of the model for environmental captioning. 
This variant utilizes Vicuna~\cite{chiang2023vicuna} as its underlying LLM, tasked with capturing and describing the operational environment of a vehicle. 
This description includes various elements such as the location and quantity of nearby vehicles and pedestrians, traffic dynamics, concerning surrounding lanes of pedestrian crossing and road.

Despite the advanced capabilities of VLMs in generating natural language descriptions, as illustrated in Figure~\ref{fig:caption}, our findings indicate that the 2D-tokenized LLM struggles with accurate environmental perception. The model frequently produces erroneous or "hallucinated" descriptions, which suggests that it still falls short of reliable perception in practical applications. This underscores the challenges and limitations inherent in deploying LLMs for complex perceptual tasks in dynamic environments.

\textbf{Remark.} 
To sum up, the experiments above reveal a significant limitation in the perception capabilities of LLMs that rely on 2D visual tokenizers. This limitation poses serious challenges for reliable ego vehicle planning. 
We claim that the primary reason for this limitation lies in the inability of 2D visual tokenizers to effectively integrate 3D spatial priors.
To address the limitation, we introduce advanced pre-trained 3D perception models as 3D tokenizers in the following section. 
\section{3D-Tokenized LLM for Reliable Autonomous Driving}

\subsection{3D-Tokenized LLM}
Distinct from 2D-tokenized LLM, we introduce 3D tokenizers founded upon a DETR-inspired architecture into LLM, formulating a 3D-tokenized LLM framework, named Atlas. 
In specific, Altas consists of three primary components. Initially, the model employs 3D tokenizers, StreamPETR~\cite{wang2023exploring}  and TopoMLP~\cite{wu2023topomlp}, to process multi-view images into DETR-style query representations. Following this, these queries are streamlined through a single linear layer, functioning as a projector, to align with the LLM. The final component of Atlas is an LLM, designed as Vicuna~\cite{chiang2023vicuna}. 
This approach brings significant benefits in \textit{incorporating 3D innate prior, achieving high resolution, and facilitating temporal propagation}, as previously elaborated.

\textbf{3D Environment Perception.}
The performance of Atlas is evaluated on standard datasets tailored to the tasks of 3D object detection and 3D lane detection, as reported in Table~\ref{tbl:detection} and Table~\ref{tbl:map}. 
The results demonstrate that 3D-tokenized LLM achieves remarkable performance across both tasks.
Besides, 3D-tokenized LLM performs better than 2D-tokenized LLM on driving environment captioning, as shown in Figure~\ref{fig:caption}, thereby affirming the significant advantages of utilizing 3D tokenizers.
In addition to representing 3D environment, our ultimate goal is to achieve reliable autonomous driving.
In the following, we will evaluate the performance of 3D-tokenized LLM on ego-car planning.



\subsection{Implementation}

The whole model trains with 8 Tesla A100 GPUs, with training times of approximately 100 hours.

\textbf{Dataset.} 
We employ the nuScenes planning dataset~\cite{caesar2020nuscenes} in our experiments of reliable autonomous driving.
As illustrated in Figure~\ref{fig:datasets},  we have reformatted the planning data into a question-answer format to facilitate our analysis.
Previous research~\cite{bevplanner} has established that the "ego states"—sensor-provided data on the autonomous vehicle such as velocity, acceleration, yaw angle, and historical trajectory—play a crucial role in open-loop planning. Additionally, to aid in navigation, especially at intersections, it is essential to incorporate a high-level command (e.g., go straight, turn left, turn right) which provides directional guidance.
Building on these insights, we propose the question-and-answer pairs demand the models to predict future velocity and acceleration based on the current state and to subsequently generate planning waypoints for the ego-car prompting by a high-level command. This processing called chain-of-thought~\cite{wei2022chain}, not only enhances the interpretability of the model's reasoning process but also its reliability. A typical example is shown in Figure~\ref{fig:datasets}, and additional details about the dataset are available in the supplementary materials.


\textbf{Metrics.} We adhere to standard practices by utilizing the implementation provided by ST-P3~\cite{stp3} to assess planning over time horizons of 1s, 2s, and 3s. We assess the performance with two widely accepted metrics: the L2 error calculated by comparing the predicted trajectories of the ego vehicle with the ground-truth trajectories at corresponding waypoints, and the collision rate calculated by checking for any intersections between the ego vehicle and other entities within the scene.

\subsection{Main Results}

\begin{table*}[t]
\caption{\textbf{Comparisons on the planning.} For a fair comparison, we refer to the reproduced results in BEV-Planner~\cite{bevplanner}. The bold \textbf{numbers} represent the highest accuracy.}
\centering
\resizebox{\linewidth}{!}{
\begin{tabular}{l|c|cc|cccc|cccc}
    \toprule
    \multicolumn{1}{l|}{\multirow{2}*{Method}} & High-level & \multicolumn{2}{c|}{Ego States} & \multicolumn{4}{c|}{L2 (m)} & \multicolumn{4}{c}{Collision (\%)} \\
    & Command & Bev & Planner & 1s & 2s & 3s & Avg. & 1s & 2s & 3s & Avg. \\
    \midrule
    FF~\cite{FF} & \ding{56} & \ding{52} & \ding{52} &0.55 & 1.20 & 2.54 & 1.43 & 0.06 & 0.17 & 1.07 & 0.43 \\
    \midrule
    \multicolumn{1}{l|}{\multirow{2}*{ST-P3~\cite{stp3}}} & \ding{52} & \ding{56} & \ding{56} & 1.59 & 2.64 & 3.73 & 2.65 & 0.69 & 3.62 & 8.39 & 4.23\\
    & \ding{52} & \ding{52} & \ding{52} & 1.33 & 2.11 & 2.90 & 2.11 & 0.23 & 0.62 & 1.27 & 0.71 \\
    \midrule
    \multicolumn{1}{l|}{\multirow{2}*{UniAD~\cite{hu2023planning}}} & \ding{52} & \ding{56} & \ding{56} & 0.59 & 1.01 & 1.48 & 1.03 & 0.16 & 0.51 & 1.64 & 0.77\\
    & \ding{52} & \ding{52} & \ding{52} & 0.20 & 0.42 & 0.75 & 0.46& 0.02 & 0.25 & 0.84 & 0.37 \\
    \midrule
    \multicolumn{1}{l|}{\multirow{2}*{VAD-Base~\cite{VAD}}} & \ding{52} & \ding{56} & \ding{56} & 0.69 & 1.22 & 1.83 & 1.25 & 0.06 & 0.68 & 2.52 & 1.09\\
    & \ding{52} & \ding{52} & \ding{52} & 0.17 & 0.34 & 0.60 & 0.37 & 0.04 & 0.27 & 0.67 & 0.33 \\
    \midrule
    Ego-MLP~\cite{admlp} & \ding{52} & \ding{56} & \ding{52} & \textbf{0.15} & 0.32 & 0.59 & 0.35 & \textbf{0.00} & 0.27 & 0.85 & 0.37 \\
    \midrule
    \multicolumn{1}{l|}{\multirow{2}*{BEV-Planner~\cite{bevplanner}}} & \ding{52} & \ding{56} & \ding{56} & 0.30 & 0.52 & 0.83 & 0.55 & 0.10 & 0.37 & 1.30 & 0.59\\
    & \ding{52} & \ding{52} & \ding{52} & 0.16 & 0.32 & 0.57 & 0.35 & \textbf{0.00} & 0.29 & 0.73 & 0.34 \\
    \midrule
    LLaVA~\cite{liu2024visual} & \ding{52} & \ding{56} & \ding{56} & 1.04 & 1.74 & 2.57 & 1.79 & 0.58 & 1.17 & 1.74 & 1.16 \\
    \midrule
    Vicuna~\cite{chiang2023vicuna} & \ding{52} & \ding{56} & \ding{56} & 1.06 & 1.80 & 2.54 & 1.80 & 0.60 & 1.21 & 1.78 & 1.20 \\
    \midrule
    Merlin~\cite{yu2023merlin} & \ding{52} & \ding{56} & \ding{56} & 1.03 & 1.71 & 2.40 & 1.71 & 0.48 & 1.05 & 1.77 & 1.10 \\
    \midrule
    \multicolumn{1}{l|}{\multirow{3}*{Atlas}} & \ding{56} & \ding{56} & \ding{56} & 1.69 & 1.89 & 2.25 & 1.94 & 0.51 & 0.85 & 1.44 & 0.93\\
    & \ding{52} & \ding{56} & \ding{56} & 0.52 & 0.97 & 1.53 & 1.00 & 0.15 & 0.31 & 0.70 & 0.38 \\
    & \cellcolor{gray!15}{\ding{52}} & \cellcolor{gray!15}{\ding{52}} & \cellcolor{gray!15}{\ding{52}} & \cellcolor{gray!15}{0.18} & \cellcolor{gray!15}\textbf{0.21} & \cellcolor{gray!15}\textbf{0.26} & \cellcolor{gray!15}\textbf{0.21} & \cellcolor{gray!15}{0.12} & \cellcolor{gray!15}\textbf{0.13} & \cellcolor{gray!15}\textbf{0.16} & \cellcolor{gray!15}\textbf{0.13} \\
    
    \bottomrule
    \end{tabular}
    }
    \vspace{-5pt}
    \label{tbl:planning}
\end{table*}

In this section, we evaluate the performance of our proposed method, Atlas, by comparing it against existing state-of-the-art (SoTA) BEV-based planners, as detailed in Table \ref{tbl:planning}. Our experimental results reveal that Atlas achieves substantial improvements over the SoTA methods, reducing the average L2 metric by 40.0\% and the average Collision metric by 60.6\%. These significant enhancements corroborate the effectiveness of the 3D-tokenized LLMs, which we consider as the key to reliable autonomous driving.

Further, to ascertain whether the performance improvements are solely attributable to the inclusion of ego state information—a frequent topic of discussion within the community—we conduct additional experiments by removing the ego state data during both training and testing.
In this experimental setting, compared to the prevailing VLM-based methods, our Atlas demonstrates superior performance and robustly validates the effectiveness of 3D tokenizers.
Despite this, Atlas continues to outperform other BEV-based methods in terms of collision rates. However, the performance on the L2 metric is comparable to other methods. We hypothesize that this outcome may stem from the inherent capabilities of the LLM to predict multiple potential motion trajectories and make rational decisions, which, while confronted with novel scenarios, deviate from the ground truth.

\subsection{Ablation Study}
\label{sec:ablation}
To avoid unnecessary misunderstandings, our ablation does not introduce any ego states.

\begin{table}[t]
	\caption{A set of ablative studies on 3D object detection and ego-car planning. The adopted algorithm designs and hyper-parameter settings are marked in \colorbox{gray!15}{\textbf{bold}}. See \S\ref{sec:ablation} for details.}
	\hspace{-0.7em}
	\begin{subtable}{0.53\linewidth}
		\captionsetup{width=.95\linewidth}
		\resizebox{\textwidth}{!}{
			\setlength\tabcolsep{10pt}
			\renewcommand\arraystretch{1.12}
			\begin{tabular}{c|cc|c c}
                \toprule
                \multicolumn{1}{l|}{\multirow{2}*{}} & \multicolumn{2}{c|}{3D Detection} & \multicolumn{2}{c}{Planning} \\
                & \textbf{F1$_{1.0}$} & \textbf{F1$_{2.0}$} & \textbf{Avg. L2} & \textbf{Avg. Col.}\\
                \midrule
                Vicuna & 5.4 & 10.1 & 2.19 & 2.75  \\
                $+$QR  & 30.7 & 41.2 & 1.22 & 0.62 \\
                $+$ RP & 34.6 & 46.5 & 1.10 & 0.44 \\
                \cellcolor{gray!15}\textbf{$+$ MQ} & \cellcolor{gray!15}\textbf{39.8} & \cellcolor{gray!15}\textbf{49.7} & \cellcolor{gray!15}\textbf{1.00} & \cellcolor{gray!15}\textbf{0.38} \\ 
                \bottomrule
			\end{tabular}
		}
		\vspace{3px}
		\setlength{\abovecaptionskip}{0.3cm}
		\setlength{\belowcaptionskip}{-0.1cm}
		\caption{\small{Component Effect. QR, RP, and MQ mean Query Representation, Reference Point embedding, and Memory Queue.}}
		\vspace{-6px}
		\label{table:cumulative}
	\end{subtable}
	\hspace{-0.5em}
	\begin{subtable}{0.45\linewidth}
		\resizebox{\textwidth}{!}{
			\setlength\tabcolsep{8pt}
			\renewcommand\arraystretch{1.2}
			\begin{tabular}{cccc|cc}
                \toprule
                 PT & SP & TM & \textbf{Avg. L2} & \textbf{Avg. Col.}\\
                \midrule
                \ding{52} & - & - & 1.51 & 1.05\\
                - & \ding{52} & - & 1.06 & 0.41\\
                 \cellcolor{gray!15}\textbf{-} & \cellcolor{gray!15}\textbf{\ding{52}} & \cellcolor{gray!15}\textbf{\ding{52}} & \cellcolor{gray!15}\textbf{1.00} & \cellcolor{gray!15}\textbf{0.38}\\
                \bottomrule
			\end{tabular}
		}
		\vspace{10px}
		\setlength{\abovecaptionskip}{0.4cm}
		\setlength{\belowcaptionskip}{-0.1cm}
		\caption{\small{Effect of different 3D tokenizers. PT, SP and TM represent PETR, StreamPETR and TopoMLP.}}
		\vspace{-6px}
		\label{table:3D tokenizers}
	\end{subtable}
	\hspace{-0.5em}
	\begin{subtable}{0.33\linewidth}
		\resizebox{\textwidth}{!}{
			\setlength\tabcolsep{4pt}
			\renewcommand\arraystretch{1.28}
			\begin{tabular}{c|cc}
                \toprule
                Resolution & \textbf{Avg. L2} & \textbf{Avg. Col.}\\
                \midrule
                336$\times$336 & 1.66 & 0.94 \\
                320$\times$800 & 1.41 & 0.58 \\
                \cellcolor{gray!15}\textbf{800$\times$1600} & \cellcolor{gray!15}\textbf{1.00} & \cellcolor{gray!15}\textbf{0.38} \\
                \bottomrule
			\end{tabular}
		}
		\vspace{-3px}
		\setlength{\abovecaptionskip}{0.8cm}
		\setlength{\belowcaptionskip}{0cm}
		\caption{\small{Effect of resolution.}}
		\vspace{12px}
		\label{table:resolution}
	\end{subtable}
	\begin{subtable}{0.33\linewidth}
		\resizebox{\textwidth}{!}{
			\setlength\tabcolsep{4pt}
			\begin{tabular}{c|cc}
                \toprule
                RP. emb. & \textbf{Avg. L2} & \textbf{Avg. Col.}\\
                \midrule
                none & 1.18 & 0.57 \\
                sin-cos & 1.21 & 0.57 \\
                learned & 1.19 & 0.56 \\
                \cellcolor{gray!15}\textbf{RP} & \cellcolor{gray!15}\textbf{1.00} & \cellcolor{gray!15}\textbf{0.38} \\
                \bottomrule
			\end{tabular}
		}
		\vspace{-2px}
		\setlength{\abovecaptionskip}{0.32cm}
		\setlength{\belowcaptionskip}{0cm}
		\caption{\small{Reference point embeddings.}}
		\label{table:position embeddings}
	\end{subtable}
	\begin{subtable}{0.33\linewidth}
		\resizebox{\textwidth}{!}{
			\setlength\tabcolsep{4pt}
			\begin{tabular}{c|cc}
            \toprule
            LLMs & \textbf{Avg. L2} & \textbf{Avg. Col.}\\
            \midrule
            LLaMA~\cite{touvron2023llama} & 1.14 & 0.47 \\
            LLaVA~\cite{liu2024visual} & 1.03  & 0.39  \\
            \cellcolor{gray!15}\textbf{Vicuna}~\cite{chiang2023vicuna} & \cellcolor{gray!15}{1.00} & \cellcolor{gray!15}\textbf{0.38}  \\
            Merlin~\cite{yu2023merlin} & \textbf{0.99}  & 0.42  \\
            \bottomrule
            \end{tabular}
		}
		\setlength{\abovecaptionskip}{0.4cm}
		\setlength{\belowcaptionskip}{-0.1cm}
		\caption{\small{Different pretrained LLMs.}}
		\label{table:LLMs}
	\end{subtable}
 \vspace{-14mm}
\end{table}

\textbf{Component Effect.} 
We conduct ablation studies to analyze our proposed 3D-tokenized LLM, considering several key aspects: \textit{query representation}, \textit{reference point embedding}, and \textit{memory queue}, all decoupling from StreamPETR on 3D detection and planning.
The results are summarized in Table \ref{table:cumulative}.
Our experiments demonstrate that each component progressively enhances performance in both tasks. Furthermore, we observe a synergistic effect where improvements in one task appear to amplify accuracy in the other, strongly proving that the capacity to perceive the environment remains a cornerstone of reliable planning.

\textbf{3D Tokenizers.} 
We investigate the effectiveness of various 3D tokenizers for ego-car planning, which are the central enhancements introduced in our study. The results are shown in Table \ref{table:3D tokenizers}. The tokenizers we evaluate include PETR (PT)~\cite{liu2022petr}, StreamPETR (SP)~\cite{wang2023exploring}, and TopoMLP (TM)~\cite{wu2023topomlp}. Our incorporation of progressively advanced 3D perceptrons into LLM demonstrates a notable improvement in planning performance, underscoring the significance of 3D perception in achieving robust autonomous driving. Furthermore, we integrate TopoMLP to provide supplementary lane line information. This addition results in a modest enhancement in performance, suggesting the potential benefits of incorporating contextual roadway features into the motion planning process.

\textbf{Resolution.} 
Our approach integrates 3D tokenizers with adjustable image resolution capabilities, which aligns well with real-world applications in autonomous driving.
As Table \ref{table:resolution} presents, we observe that increasing the image resolution leads to a noticeable improvement in performance. This evidence indicates that our method holds significant advantages over traditional VLM techniques, particularly in terms of flexibility and efficacy in handling diverse image resolutions.

\textbf{Reference Point Embeddings.}
Our Atlas introduces an important concept: 3D tokenizers equipped with reference point embeddings, following the setting of StreamPETR~\cite{wang2023exploring} and TopoMLP~\cite{wu2023topomlp}. 
Here, we evaluate the model performance of decoupling reference point embedding and query embedding.
Our initial approaches relied solely on query representations (i.e., "none" in Table \ref{table:position embeddings}), which overlooks the crucial 3D spatial context—termed as the reference point. However, as shown in Table \ref{table:position embeddings}, simply applying conventional embedding techniques~\cite{carion2020end}, like sin-cos position embedding and learned position embedding, to 3D queries do not markedly influence performance. This outcome underscores the unique advantages of reference points.
To effectively utilize this, we incorporate offset mappings from the reference points via a single layer projector aka reference point embeddings to the 3D query representation (i.e., "RP" in Table \ref{table:position embeddings}).  Notably, this method achieves remarkable improvements in accuracy, underscoring its effectiveness.

\textbf{Pretrained LLMs.} 
In our experiments, we evaluate different LLMs that varied in their pre-training methodologies, as detailed in Table \ref{table:LLMs}. Our results show that LLMs pre-trained with methods that align text and images significantly outperform others in planning tasks. We attribute this enhanced performance to the multimodal nature of their training.
Additionally, our analysis reveals that models pre-trained with various visual-language data exhibited no significant differences in planning performance. We believe this is due to the absence of 3D data in their pre-training processes, suggesting that the inclusion of 3D data in pre-training, as 3D tokenizers do, is necessary.

\setlength\intextsep{6pt}
\begin{wraptable}{h}{0.3\linewidth}{
\footnotesize
\centering
\vspace{-2mm}
\vspace{1mm}
\caption{Effective of the chain of thought.
}
	   \setlength{\abovecaptionskip}{0cm}
	   \setlength{\belowcaptionskip}{0.1cm}
         \centering
  	\resizebox{0.3\textwidth}{!}{
		\setlength\tabcolsep{4pt}
		\renewcommand\arraystretch{1}
		\begin{tabular}{c|cc}
    \toprule
    chain & \textbf{Avg. L2} & \textbf{Avg. Col.}\\
    \midrule
    P & 1.33 & 0.79\\
    V-P & 1.21  & 0.61 \\
    \cellcolor{gray!15}\textbf{V-A-P} & \cellcolor{gray!15}\textbf{1.00} & \cellcolor{gray!15}\textbf{0.38} \\
    V-A-Y-P & 1.15 & 0.55 \\
    V-A-T-P & 1.40 & 0.81 \\
    P-V-A & 1.01 & 0.40 \\
    \bottomrule
    \end{tabular}}

\label{table:cot}
}
\end{wraptable}

\textbf{Chain of Thought.} 
In the realm of autonomous driving, recent works~\cite{admlp, bevplanner} converge on a key insight: the state of the ego vehicle is a pivotal factor in shaping open-loop planning strategies. 
To this end, we delineate the ego states into four distinct yet interrelated dimensions: velocity (V), acceleration (A), yaw angle (Y), and the historical trajectory (T).
To evaluate the influence of each dimension, we conduct ego planning based on the predicated of these parameters. The experimental findings, as outlined in Table~\ref{table:cot}, where "P" denotes "Planning". Notably, our results diverge from prevailing research, indicating that the yaw angle and historical trajectory do not enhance the efficacy of the planning process. This counterintuitive outcome is likely a consequence of the inherent difficulties in the precise forecasting of these variables~\cite {wei2023autoregressive, Bai_2024_CVPR}.
Moreover, we discover an interesting aspect of our model's robustness: the sequence in which these parameters are predicted does not impact the performance. This suggests that altering the order of prediction (e.g., reversed) does not increase computational time.

\subsection{Qualitative Results}

\begin{figure}[t]
    \centering
    \includegraphics[width=\linewidth]{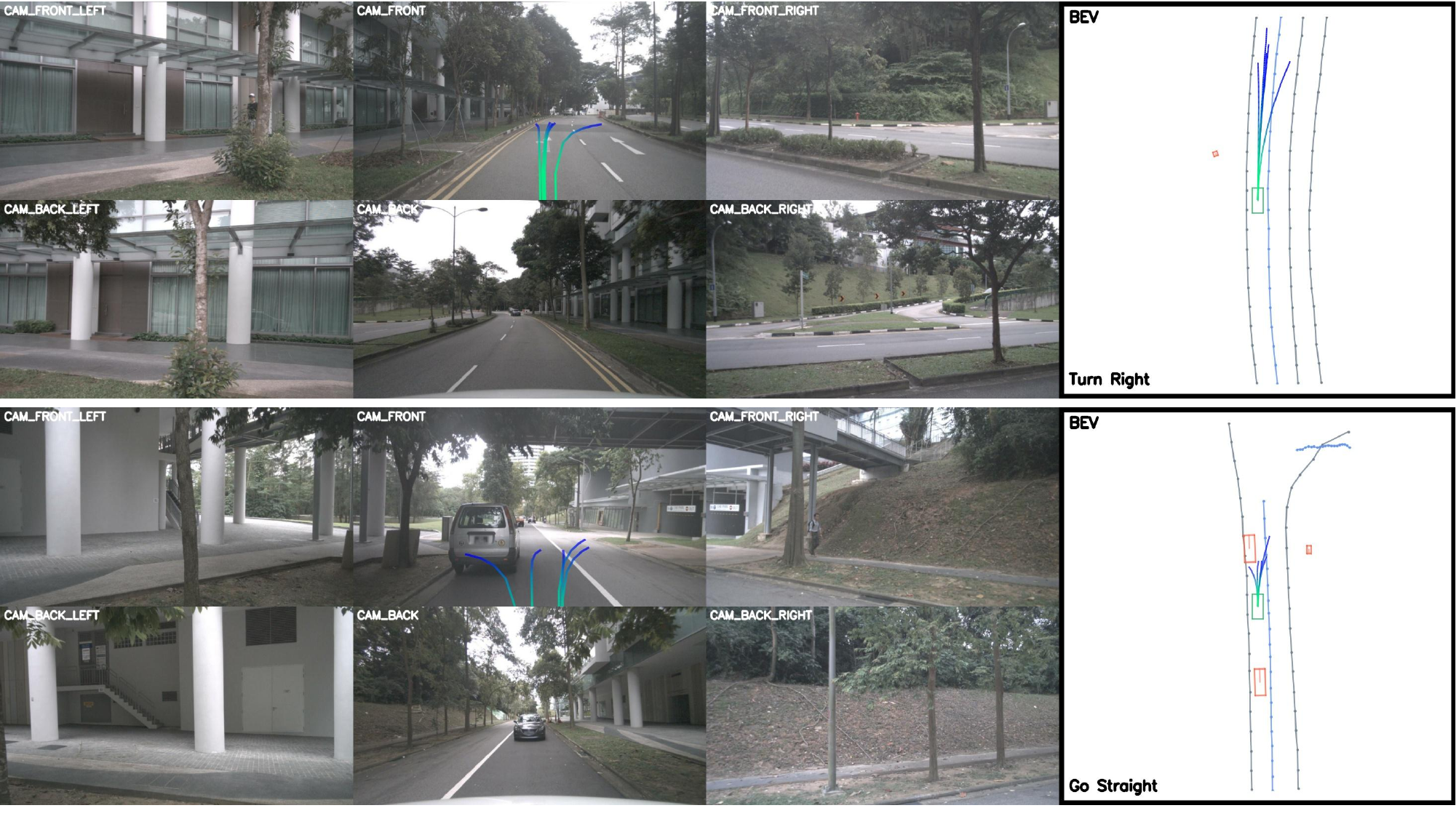}
    \vspace{-16pt}
    \caption{\textbf{Qualitative results with diverse planning from Atlas.}
    The five planning trajectories presented here are generated through five iterations of utilizing our 3D-tokenized LLM.
    It is obvious that Atlas is able to output different potential planning trajectories thanks  to LLM.}
    \label{fig:quali}
\end{figure}

We also conduct a qualitative analysis by visualizing the trajectory predictions made by Atlas, as shown in Figure \ref{fig:quali}. 
We execute the 3D-tokenized LLM five times to produce five depicted planning trajectories.
The results demonstrate that Atlas is capable of generating multiple feasible plans for autonomous driving that are not only practical but also adhere to safety standards. Specifically, Atlas successfully devises various potential routes tailored to distinct driving scenarios, including following other vehicles, lane changing, and overtaking. Importantly, Atlas effectively identifies and avoids pedestrians and cars, showcasing its robust capability in ensuring road safety.


\section{Related Work}
\textbf{DETR-style BEV Perception.}
DETR~\cite{carion2020end} is initially proposed to address the challenge of end-to-end detection, and further extensively applied in BEV perception~\cite{liu2022petr,li2022bevformer,liu2023sparsebev,lin2022sparse4d}, thereby significantly advancing its development. DETR3D~\cite{wang2022detr3d} is a pioneering work that introduces the concept of \textit{3D object queries}, which interact with multi-view image features to produce sparse yet informative object representations. Further, PETR~\cite{liu2022petr,liu2022petrv2} introduces the concept of 3D position encoding, and BEVFormer~\cite{li2022bevformer} brings BEV temporal modeling.
StreamPETR~\cite{wang2023exploring} and Sparsev2~\cite{lin2023sparse4d} use object queries as a vessel for temporal modeling, effectively propagating temporal information while achieving SoTA performance with commendable efficiency. 
\textit{In a notable finding within StreamPETR, the inclusion of additional multi-frame image feature interactions does not enhance performance, suggesting that the highly compressed object queries are sufficiently expressive to encapsulate all necessary information for BEV perception.}
Moreover, the application of DETR framework has been expanded to \textit{map queries} by works such as MapTR~\cite{liao2022maptr}, TopoNet~\cite{li2023topology} and TopoMLP~\cite{wu2023topomlp}, which are instrumental in the construction of vectorized map representations.

\textbf{BEV-based End-to-end Driving.}
Traditional autonomous driving systems have often relied on manual rules for planning, which can be cumbersome and complex, struggling to cover the numerous corner cases. In recent years, there has been a pronounced shift towards end-to-end autonomous driving approaches, which have demonstrated significant progress in simplifying and streamlining the autonomous driving pipeline. UniAD~\cite{hu2023planning} is a pioneering work that introduces an end-to-end framework encompassing tasks such as perception, prediction, and planning, with these tasks executed sequentially to ultimately produce control outputs. Building upon this framework, VAD~\cite{alexanian1990vad} further streamlines the pipeline, enhancing efficiency and reducing complexity. However, AD-MLP~\cite{zhai2023rethinking} and BEV-Planner~\cite{li2023ego} have observed that existing end-to-end methods can achieve high performance on open-loop benchmarks like nuScenes~\cite{caesar2020nuscenes} by simply fitting to the ego status of the autonomous vehicle. This finding suggests that the integration of planning and control in these models may not yet fully capture the complexities of real-world driving scenarios. Subsequent works, such as Think-Twice~\cite{jia2023think} and VADv2~\cite{chen2024vadv2}, have made substantial advancements in closed-loop simulators like Carla~\cite{dosovitskiy2017carla}. Following BEV-Planner~\cite{li2023ego}, we present results with and without the ego status to address the open-loop challenges on the nuScenes~\cite{caesar2020nuscenes} benchmark.

\textbf{VLM-Agent for Autonomous Driving.}
The visual-language model (VLM) domenstrates promising results in the fields of visual-language understanding and logical reasoning, and has been extended to autonomous driving ~\cite{xu2023drivegpt4,tian2024drivevlm,wang2023drivemlm,shao2023lmdrive,jia2023adriver, xie2023sed}. DriveGPT4~\cite{xu2023drivegpt4} employs a VLM model to predict driving commands and provide rational explanations for its decisions. DriveLM~\cite{sima2023drivelm} excels at conversing about environmental information, while ADriver-I~\cite{jia2023adriver} focuses on predicting low-level vehicle signals. Furthermore, DriveMLM~\cite{wang2023drivemlm} and LMDrive~\cite{shao2023lmdrive} have implemented end-to-end autonomous driving solutions and validated effectiveness on CARLA~\cite{dosovitskiy2017carla} closed-loop benchmarks, showcasing the potential of VLM-based agents.
Despite impressive progress, no work explores how the 3D-tokenized LLM influences real-life autonomous driving.


\label{sec:related}

\section{Conclusion and Limitations} 

In this paper, we explored VLM-based methods increasingly used in autonomous driving, focusing first on perception. We found large gaps between task-specific and 2D tokenized LLM-based methods in environmental perception, which is essential for reliable autonomous driving.
To address these gaps, we introduced Atlas, a system combining DETR-style 3D perceptrons with LLMs. This approach integrates 3D priors for better depth perception and supports high-resolution, multi-view images, and temporal modeling through query propagation.
Our evaluation of Atlas on nuScenes dataset revealed substantial improvements in 3D detection and planning, surpassing established methods. This confirms our belief that 3D-tokenized LLM is the key to reliable autonomous driving.

\textbf{Limitations.} This paper aims to demonstrate the effectiveness of the 3D tokenizer for VLM-based autonomous driving. Although our method has demonstrated outstanding performance in open-loop planning, it has not yet been tested on a closed-loop dataset. However, existing close-loop benchmarks (e.g., CARLA~\cite{dosovitskiy2017carla}) lack reality, which fails to verify our motivation. Moreover, this paper lacks of performance comparison with VLM-based AD methods. This omission is due to the proprietary codes for these methods.


\label{sec:conclusion}


{\small
\bibliographystyle{unsrt}
\bibliography{neurips_2024}
}

\newpage
\appendix
\text{\LARGE\textbf{Appendix}}

\section{Datasets Details}

To align with the requirements of VLM-based models, the necessary step is to transform all evaluation datasets into a textual format, specifically structured as question-answer pairs.
In this section, we will delve deeper into the specifics of various datasets, including 3D object detection (\S \ref{sec:det}), 3D lane detection  (\S \ref{sec:lane}), driving captioning  (\S \ref{sec:caption}), and ego planning  (\S \ref{sec:planning}).

\begin{figure}[h]
    \centering
    \includegraphics[width=0.925\linewidth]{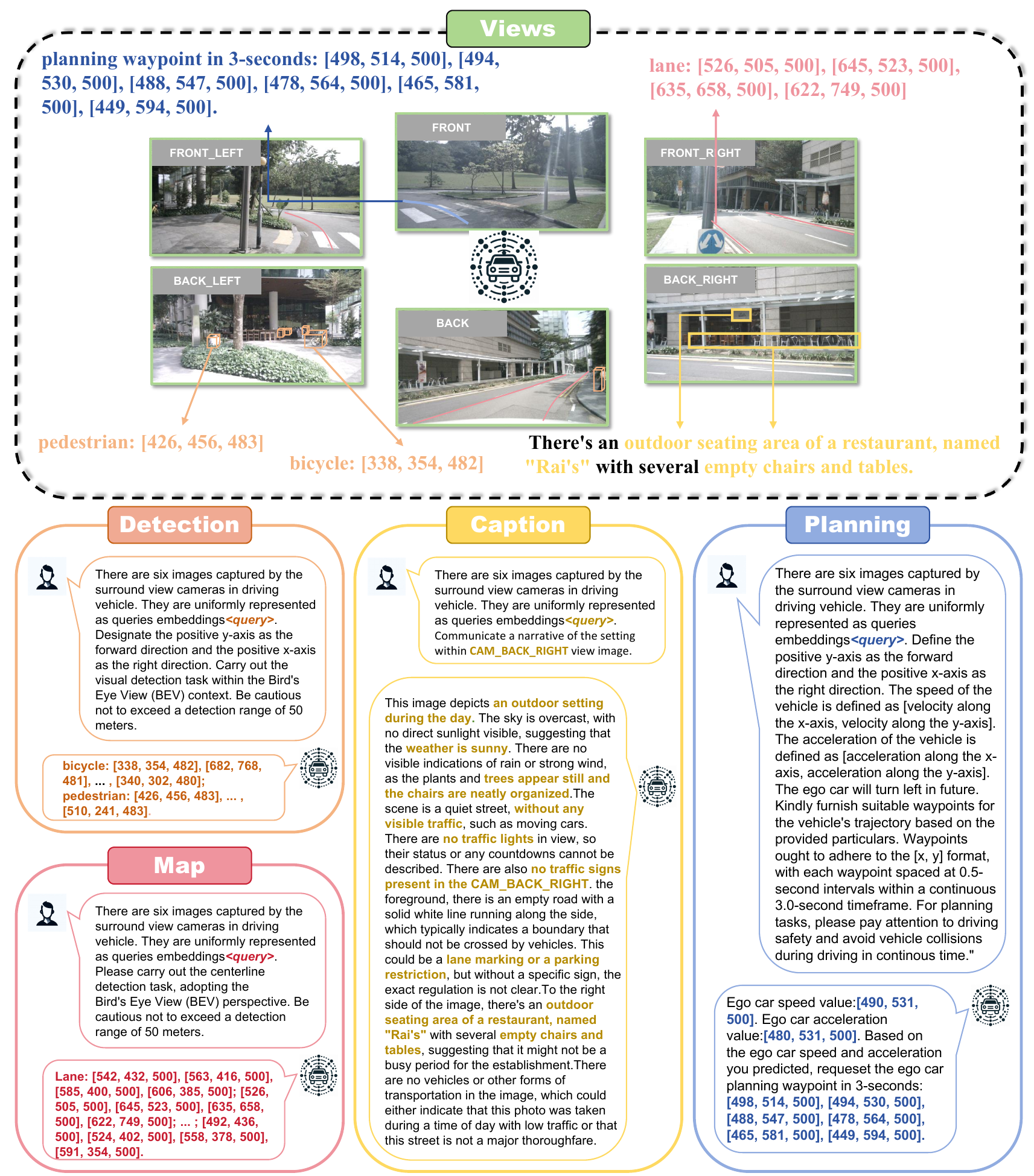}
    \vspace{-4pt}
    \caption{\textbf{Our constructed question-answer pairs for VLM-based methods.} It transforms several critical driving reasoning tasks, such as 3D object detection, map perception, environment caption, and ego-car planning, into a uniform text format.}
    \label{fig:dataset_details}
\end{figure}

\subsection{3D Object Detection}
\label{sec:det}
The 3D object detection utilized in the VLM-based method (2D/3D visual tokenizers with LLM) evaluation is based on nuScenes~\cite{caesar2020nuscenes}.
To adapt to the inputs and outputs of LLM, we convert the detection task into a text-format question-answer task.
Here, the question is randomly sampled from a pool that is listed in Table~\ref{table:detection_question}.
As seen, we set a special token `<query>' to accept tokens from 3D tokenizers.
If the inputs are six-view images, we replace the text `They are uniformly represented as queries embeddings<query>' in question with `They represent left rear image<query>, left front image<query>, direct front image<query>, right front image<query>, right rear image<query>, and direct rear image<query>.'.
As for the answer, we choose the category name and 3D center points of each bounding box, as shown in Figure~\ref{fig:dataset_details}.
To facilitate more efficient localization, we discretize the bird's-eye view (BEV) space ranging from -50 meters to +50 meters into 1,000 bins. 

\begin{table*}[ht!]\centering
\begin{minipage}{0.99\columnwidth}\vspace{0mm}    \centering
\begin{tcolorbox} 
    \centering
    \small
     \hspace{-6mm}
\begin{itemize}[leftmargin=4mm]
\setlength{\itemsep}{2pt}
\item "There are six images captured by the surround view cameras in driving vehicle. They are uniformly represented as queries embeddings<query>. Define the positive y-axis as the forward direction and the positive x-axis as the right direction. Please complete the visual detection task under the Bird's Eye View (BEV) perspective. Ensure that the detection range does not exceed 50 meters."
\item "There are six images captured by the surround view cameras in driving vehicle. They are uniformly represented as queries embeddings<query>.  Establish the positive y-axis as the frontward direction and the positive x-axis as the rightward direction. Kindly execute the visual detection task within the Bird's Eye View (BEV) framework. Be mindful not to exceed a detection range of 50 meters."
\item "There are six images captured by the surround view cameras in driving vehicle. They are uniformly represented as queries embeddings<query>. Set the forward direction as the positive y-axis and the right direction as the positive x-axis. Please carry out the visual detection task within the Bird's Eye View (BEV) context. Ensure that the detection range remains within 50 meters."
\item  \dots  \dots
\end{itemize}
\end{tcolorbox}
\vspace{-2mm}
\caption{\textbf{Question pool of 3D object detection for VLM-based methods.}}
    \label{table:detection_question}
\end{minipage}
\end{table*}

\subsection{3D Lane Detection}
\label{sec:lane}

We formulate a 3D lane detection dataset with question-answer pairs based on the OpenLane-V2 Subset-B~\cite{wang2024openlane}, which itself originates from the nuScenes dataset. 
A representative is shown in Figure~\ref{fig:dataset_details}.
Their questions are sampled from Table~\ref{table:map_question}, and the corresponding answer comprises a set of four lane points. Analogous to the 3D object detection dataset, we discretize the BEV space, spanning from -50 meters to +50 meters, into 1,000 bins.

\begin{table*}[ht!]\centering
\begin{minipage}{0.99\columnwidth}\vspace{0mm}    \centering
\begin{tcolorbox} 
    \centering
    \small
     \hspace{-6mm}
\begin{itemize}[leftmargin=4mm]
\setlength{\itemsep}{2pt}
\item "There are six images captured by the surround view cameras in driving vehicle. They are uniformly represented as queries embeddings<query>. Please complete the centerline detection task under the Bird's Eye View (BEV) perspective. Ensure that the detection range does not exceed 50 meters."
\item "There are six images captured by the surround view cameras in driving vehicle. They are uniformly represented as queries embeddings<query>.   Be mindful not to exceed a detection range of 50 meters."
\item "There are six images captured by the surround view cameras in driving vehicle. They are uniformly represented as queries embeddings<query>. Could you complete the task of detecting the centerline from the Bird's Eye View (BEV) perspective? Ensure that the detection range remains within 50 meters."
\item \dots  \dots
\end{itemize}
\end{tcolorbox}
\vspace{-2mm}
\caption{\textbf{Question pool of 3D lane detection for VLM-based methods.}}
    \label{table:map_question}
\end{minipage}
\end{table*}

\subsection{Driving Captioning}
\label{sec:caption}
Our driving captioning dataset is created through the annotation of nuScenes, leveraging the capabilities of GPT-4V. The specific prompt utilized in GPT-4V is detailed in Table~\ref{table:prompt_gpt4v}, while an illustrative example is presented in Figure~\ref{fig:dataset_details}. It is worth mentioning that, to harness the full potential of GPT-4V, we request a unique description for each individual view, resulting in a total of approximately 180k question-answer pairs.

\begin{table*}[ht!]\centering
\begin{minipage}{0.99\columnwidth}\vspace{0mm}    \centering
\begin{tcolorbox} 
    \centering
    \small
     \hspace{-6mm}
\begin{itemize}[leftmargin=4mm]
\setlength{\itemsep}{2pt}
\item "Describe the current traffic conditions. If there are traffic lights in the image, describe the status of all the traffic lights, including any countdowns; if there are none, please do not respond.  If there are traffic signs in the picture, identify and explain each one; if there are none, no explanation is necessary. If there are other vehicles in the picture, describe them in more detail.  Please ensure the answer does not exceed 600 words. Answers must be in English."
\end{itemize}
\end{tcolorbox}
\vspace{-2mm}
\caption{\textbf{Prompt used in GPT-4V for caption generation.}}
    \label{table:prompt_gpt4v}
\end{minipage}
\end{table*}

\subsection{Ego Planning}
\label{sec:planning}
Similar to 3D object and lane detection, we adapt the nuScenes dataset into a question-answer pairs format. Following the chain-of-thought approach, we prompt our model to generate safe driving plans and to describe various ego states, such as velocity and acceleration. The specific questions used are sampled from Table \ref{table:planning_question}. For the answers, the model predicts the current state's velocity and acceleration and then generates the ego-car's planning waypoints for the next 3 seconds at 0.5-second intervals. This approach mirrors our methods in 3D object detection and 3D lane detection, where we discretize the BEV space, which ranges from -50 to +50 meters, into 1,000 bins. Similarly, we discretize both velocity and acceleration across a range from -50 m/s (m/s$^2$) to +50 m/s (m/s$^2$) into 1,000 bins each.

\begin{table*}[ht!]\centering
\begin{minipage}{0.99\columnwidth}\vspace{0mm}    \centering
\begin{tcolorbox} 
    \centering
    \small
     \hspace{-6mm}
\begin{itemize}[leftmargin=4mm]
\setlength{\itemsep}{2pt}
\item "The six images include objects that are uniformly represented as 3D detection query embeddings<query> and map query embeddings<query>. Define the positive y-axis as the forward direction and the positive x-axis as the right direction.
The speed of the vehicle is defined as [velocity along the x-axis, velocity along the y-axis]. 
The acceleration of the vehicle is defined as [acceleration along the x-axis, acceleration along the y-axis].
The ego car will turn left in future.
Kindly furnish suitable waypoints for the vehicle's trajectory based on the provided particulars. Waypoints ought to adhere to the [x, y] format, with each waypoint spaced at 0.5-second intervals within a continuous 3.0-second timeframe.
For planning tasks, please pay attention to driving safety and avoid vehicle collisions during driving in continous time.
"
\item "The six images include objects that are uniformly represented as 3D detection query embeddings<query> and map query embeddings<query>. Define the positive y-axis as the forward direction and the positive x-axis as the right direction.
The speed of the vehicle is defined as [velocity along the x-axis, velocity along the y-axis]. 
The acceleration of the vehicle is defined as [acceleration along the x-axis, acceleration along the y-axis].
The ego car will turn right in future.
We request your provision of pertinent waypoints for the vehicle's route in accordance with the given information. Waypoints should conform to the format [x, y], with spacing set at 0.5-second intervals over a continuous duration of 3.0 seconds.
For planning tasks, please pay attention to driving safety and avoid vehicle collisions during driving in continous time.
"
\item "The six images include objects that are uniformly represented as 3D detection query embeddings<query> and map query embeddings<query>. Define the positive y-axis as the forward direction and the positive x-axis as the right direction.
The speed of the vehicle is defined as [velocity along the x-axis, velocity along the y-axis]. 
The acceleration of the vehicle is defined as [acceleration along the x-axis, acceleration along the y-axis].
The ego car will go stright in future.
Please submit fitting waypoints for the vehicle's course based on the supplied data. Ensure waypoints are structured as [x, y] and spaced at intervals of 0.5 seconds across a continuous 3.0-second period.
For planning tasks, please pay attention to driving safety and avoid vehicle collisions during driving in continous time.
"
\item \dots  \dots
\end{itemize}
\end{tcolorbox}
\vspace{-2mm}
\caption{\textbf{Question pool of ego planning for VLM-based methods.}}
    \label{table:planning_question}
\end{minipage}
\end{table*}

\section{Model Details}

\subsection{3D Tokenizers Pre-training}
We pre-train two distinct 3D tokenizers: StreamPETR~\cite{wang2023exploring} and TopoMLP~\cite{wu2023topomlp}.
StreamPETR~\cite{wang2023exploring} is designed for multi-view 3D object detection. We utilize a ViT-L backbone~\cite{eva02} and process images at a high resolution of 800x1600. Moreover, we follow the official training schedule established for the nuScenes dataset.
TopoMLP~\cite{wu2023topomlp} focuses on constructing vectorized maps from multiple views. To maintain methodological consistency with StreamPETR, we employ the same ViT-L backbone and resolution. The training strategy for TopoMLP also mirrors the official.

\subsection{3D-tokenized LLM}

\textbf{Query Representation.} For the innate priors of the 3D physical world, the query-based BEV framework is introduced. These DETR-style methods, StreamPETR and TopoMLP, extract target-aware query embeddings aka query representations (content) with reference points (localization) to represent objects from multi-view images.

\textbf{Reference Point Embeddings.} 
As previously mentioned, a target is characterized by both its content and location. We integrate the query embeddings by adding reference point embeddings, which are generated from reference points via a single linear layer, to formulate the 3D tokens that represent target information. \textit{A notable aspect of our setup is we initialize the weight of the reference point projector to zero.}

\textbf{Memory Queue.} 
Taking inspiration from StreamPETR, our approach involves the storage of historical queries to preserve continuity in time, as memory queues. Specifically, we concatenate these memory queries with current queries for temporal modeling. 
To elaborate, our method includes storing queries from three additional frames that exhibit the highest confidence—specifically, the top-K queries, where in our implementation, K is set to 256. The management of these queues adheres to a first-in, first-out (FIFO) principle. 

Our 3D-tokenized LLM, Atlas, integrates the 3D tokenizers described earlier with an LLM, specifically the Vicuna-1.5. This LLM has been pre-trained on a diverse open-world data corpus, providing a robust foundation for understanding and processing spatial-temporal. Atlas follows most of the basic settings in Merlin, with a batch size of 1, a learning rate of 2e-5, and the AdamW optimizer with a weight decay of 1e-4. We implement a linear warm-up phase consisting of the first 3\% steps in total. Following the warm-up, we transition to a cosine learning rate strategy. The maximum length of prompt tokens is 4096.

\section{3D Detection Results}

\begin{figure}[t]
    \centering
    \includegraphics[width=0.925\linewidth]{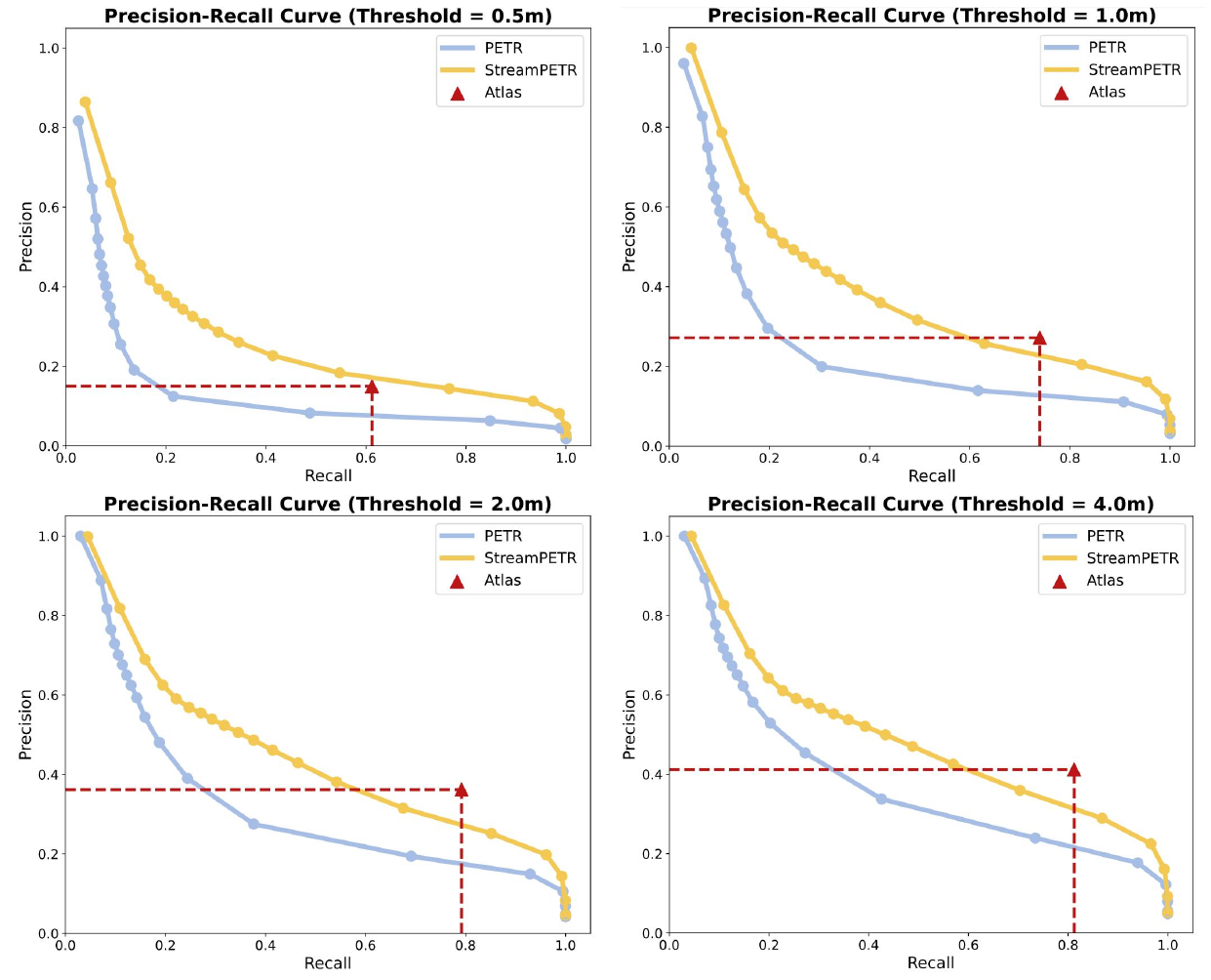}
    \vspace{-4pt}
    \caption{\textbf{Compare of 3D detection with various thresholds.} We provide the precision-recall curves of PETR and StreamPETR. As the predictions from Atlas do not include confidence scores, we calculate the precision and recall across all predicted samples.
}
    \label{fig:pr-curve}
\end{figure}

\textbf{Precion-Recall Curve.} In the paper text, we present a comparison of the F1 scores between task-specific models and Atlas in 3D detection, focusing on predictions with a confidence score above $0.3$, which yielded the highest F1 score. Additionally, we illustrate the performance variations of PETR, StreamPETR, and Atlas through the Precision-Recall curves at different positive thresholds, as shown in Figure \ref{fig:pr-curve}.
It's important to note that Atlas does not generate confidence scores; therefore, we treated all its predictions as positive samples for the purpose of calculating precision and recall. Although Atlas shows slightly weaker performance in making fine-grained predictions (specifically at a threshold of 0.5 meters), it excels in scenarios with larger thresholds. This observation suggests that large language models like Atlas might struggle with highly precise numerical predictions but perform well when broader tolerances are acceptable.

\section{More Qualitative Results}

\subsection{Qualitative Results of 3D Detection}

\begin{figure}[htbp]
    \centering
    \includegraphics[width=\linewidth]{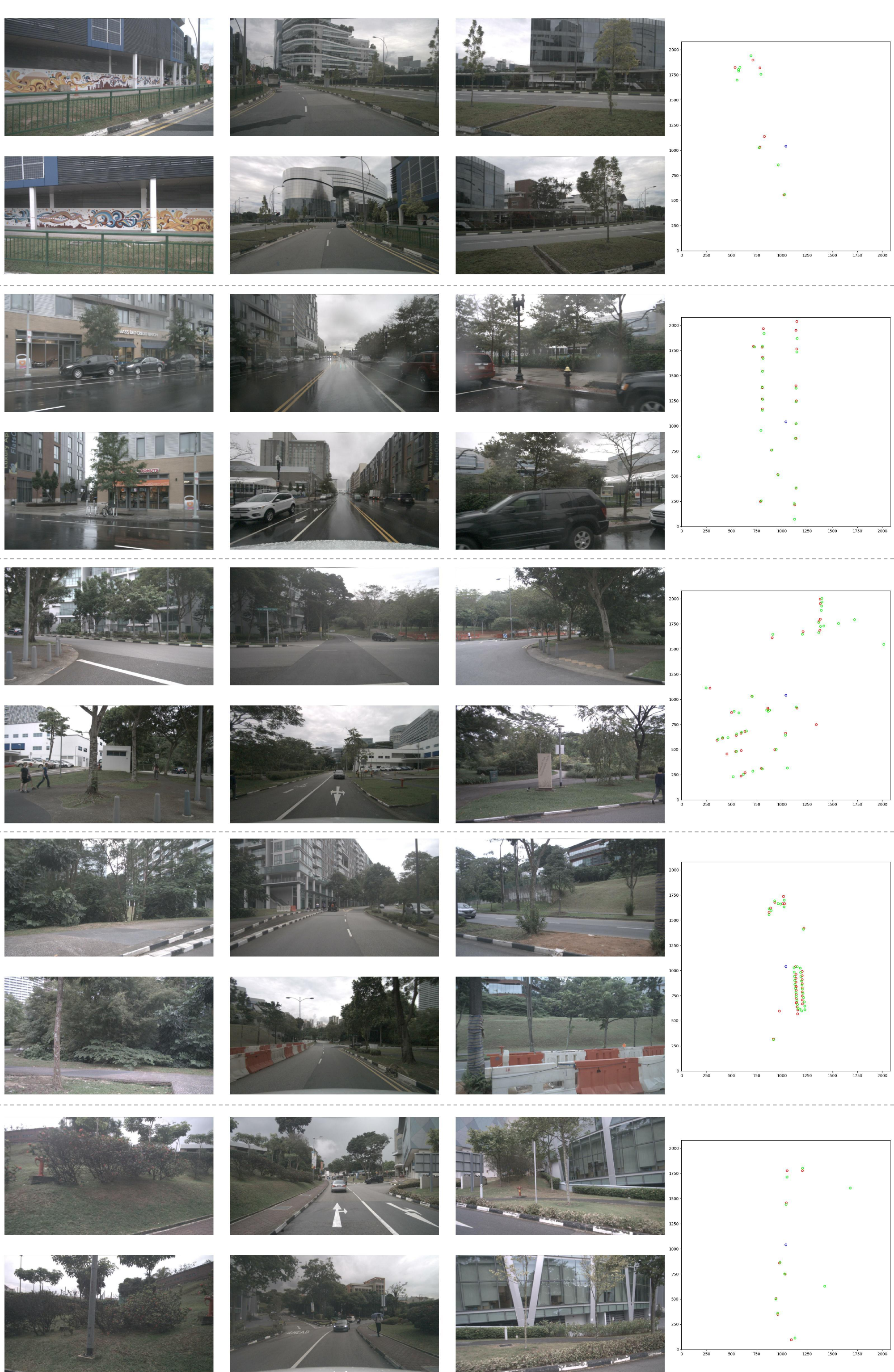}
    \vspace{-4mm}
    \caption{\textbf{Qualitative results of Atlas on 3D object detection.} The \textcolor{red}{red} circles represent the predicted objects and \textcolor{green}{green} circles represent the ground truth.
}
    \label{fig:quali-detection}
\end{figure}

We visualize the prediction results of the Atlas model in 3D detection tasks, as shown in Figure~\ref{fig:quali-detection}. The results align well with our performance metrics, demonstrating a notably high recall rate. This high recall is particularly important in practical applications of autonomous driving, where accurately detecting every potential obstacle, like pedestrians, is critical. Furthermore, the model maintains its accuracy even in complex scenarios characterized by high pedestrian density or closely packed targets.
Moreover, Atlas proves robust under challenging environmental conditions. For instance, even on rainy days, the model continues to perform strongly. This resilience is essential for the reliability needed in real-world applications, ensuring consistent performance regardless of weather conditions.

\subsection{Qualitative Results of 3D Lane Detection}

\begin{figure}[htbp]
    \centering
    \includegraphics[width=\linewidth]{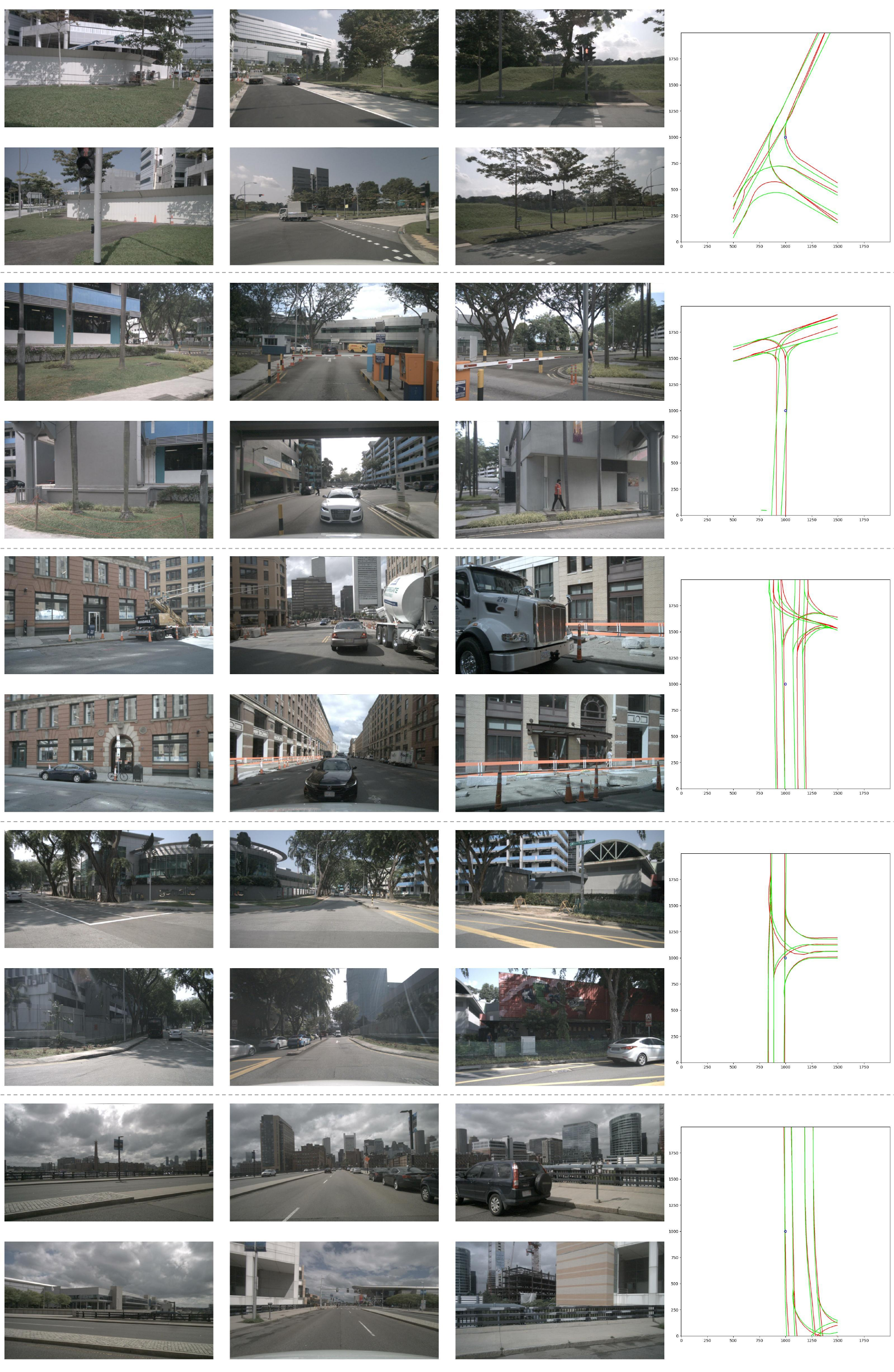}
    \vspace{-4mm}
    \caption{\textbf{Qualitative results of Atlas on map detection.} The \textcolor{red}{red} represent the predicted lane and \textcolor{green}{green} represent the ground truth.
}
    \label{fig:quali-map}
\end{figure}

We showcase the visualization outcomes of Atlas in its application to 3D lane detection, depicted in Figure~\ref{fig:quali-map}. While the quantitative performance does not surpass task-specific models, Atlas demonstrates noteworthy qualitative performance.
As seen, our model performs well in challenging road situations because it accurately recognizes road crossings and dividings.

\subsection{Qualitative Results of Planning}

\begin{figure}[htbp]
    \centering
    \includegraphics[width=\linewidth]{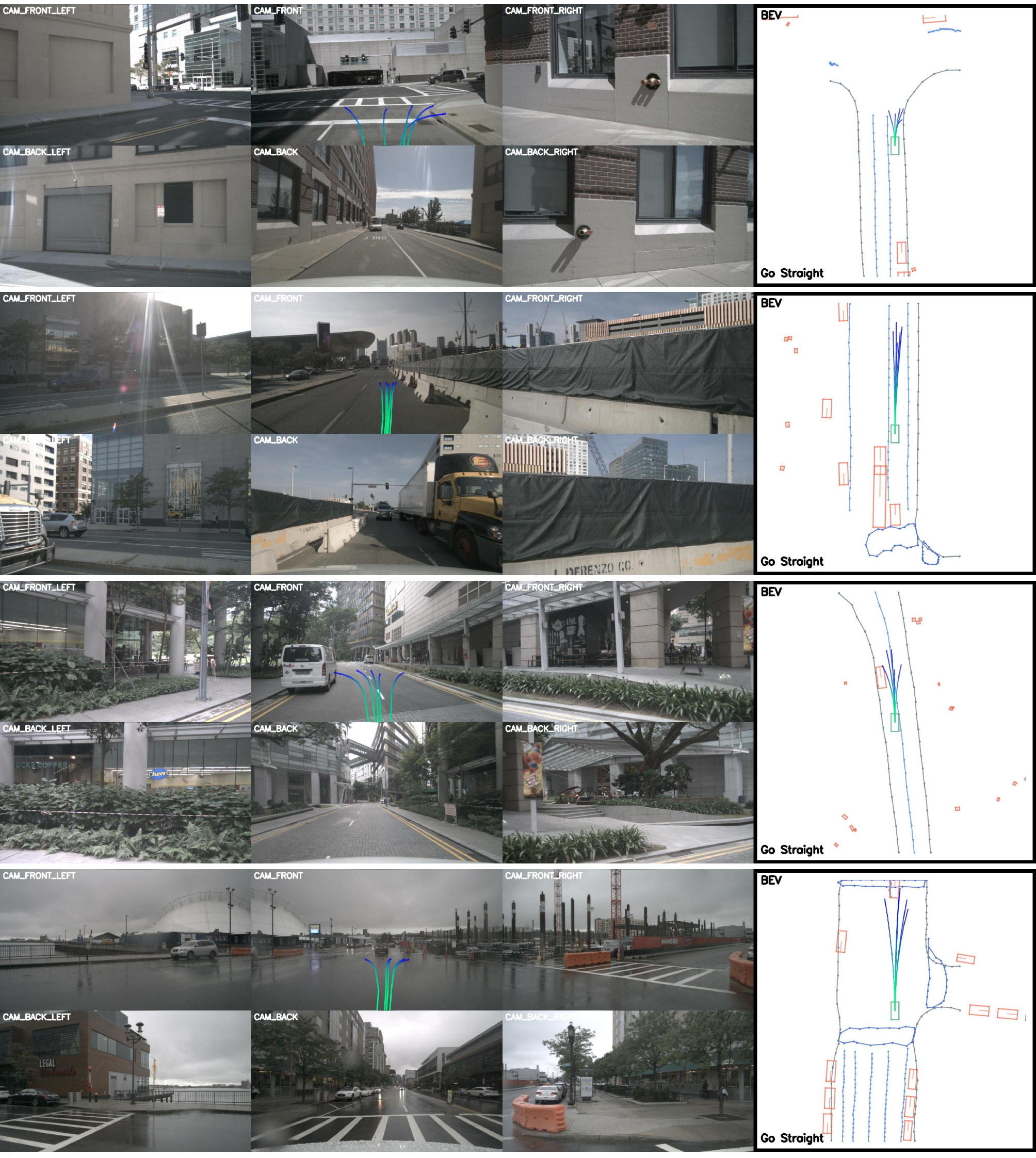}
    \vspace{-16pt}
    \caption{\textbf{Qualitative results of Atlas on ego-car planning.} Atlas outputs multiple potential planning trajectories within diverse weather and scenarios.}
    \label{fig:more_quali}
\end{figure}

We also demonstrated the adaptability of Atlas's driving plans across various weather conditions in Figure \ref{fig:more_quali}. Notably, even during rain, Atlas effectively plans its future travel trajectories with considerable diversity. This capability underscores the model's robustness in challenging environments.
Furthermore, Atlas impressively maintains compliance with traffic signals, such as stopping at red lights, without having undergone specific training for traffic light recognition. This aspect highlights the model's inherent understanding and application of world knowledge relying on LLM.
Additionally, the model's diverse planning strategy enables it to effectively balance the decisions between maintaining its current lane and executing lane changes for overtaking. This flexibility greatly enhances the variety of possible travel routes, adapting dynamically to the flow of traffic and road conditions.

\section{Failure Cases}

\begin{figure}[htbp]
\centering
{
		\includegraphics[scale=0.4]{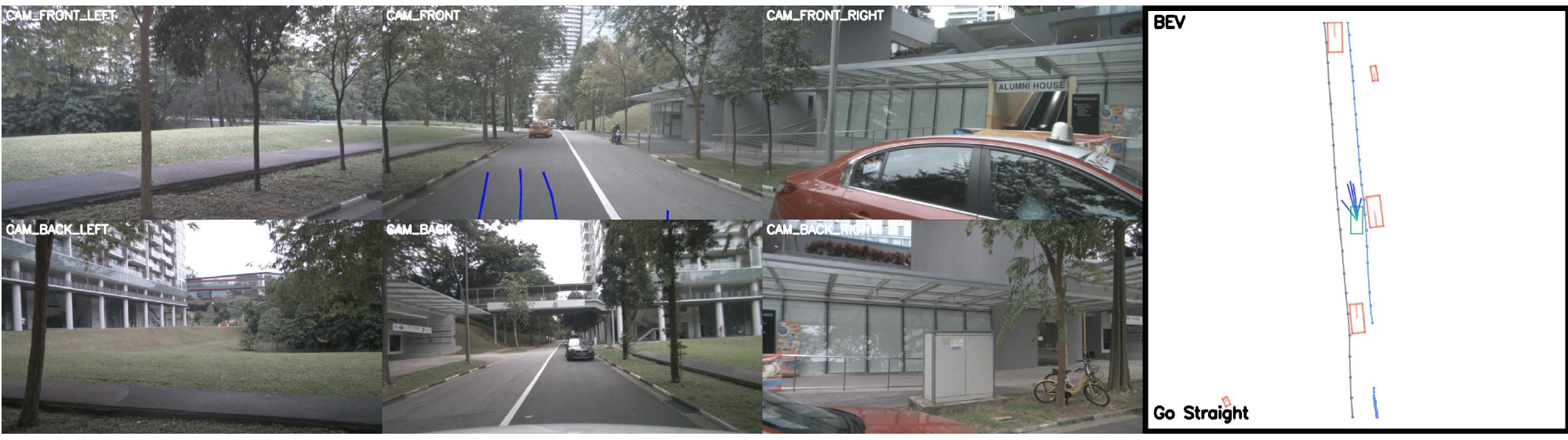}
            \caption{Overly conservative}
        \label{fig:conservative}
        }
{
		\includegraphics[scale=0.4]{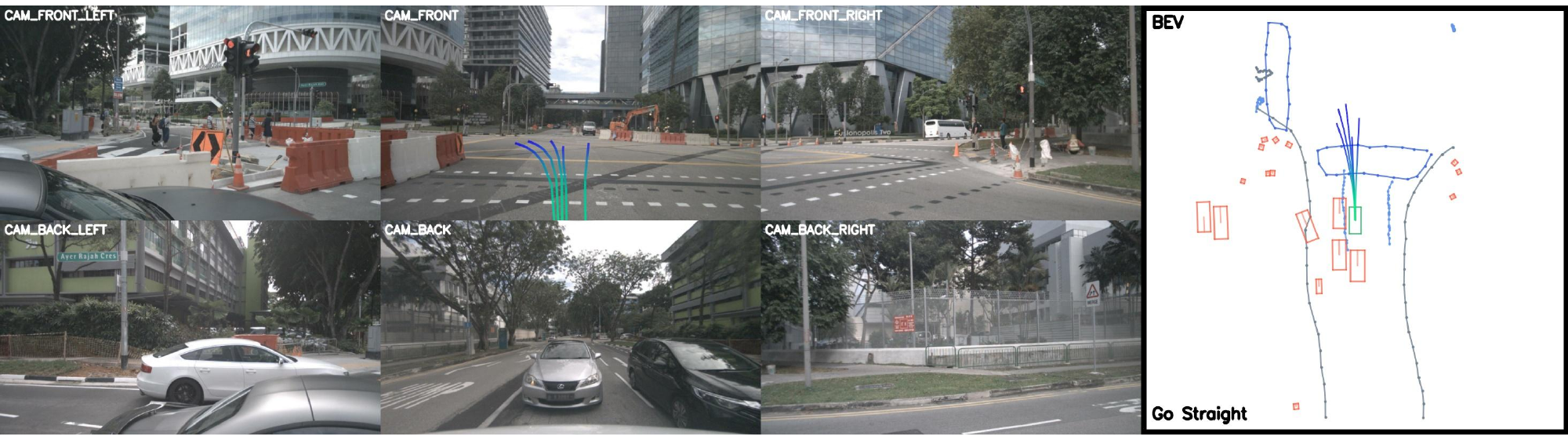}
          \caption{Violation of traffic regulations}
        \label{fig:Violation}
        }
\label{fig:fail}
\end{figure}

Discussing error examples in our model, Atlas, provides valuable insights that can guide future improvements. In this section, we analyze two primary types of failure observed during our experiments:

\textbf{Overly Conservative Behavior. }Atlas tends to make overly conservative decisions, favoring caution even when the path ahead is clear, as shown in Figure~\ref{fig:conservative}. This behavior results in a lower travel efficiency as the model opts to prioritize safety excessively. Our analysis suggests that this conservatism is likely rooted in the sampling bias of the nuScenes dataset. This dataset predominantly includes safer driving examples and favors lower-speed scenarios, which may have influenced Atlas' decision-making strategy. To address this issue, incorporating a substantial amount of closed-loop data could be beneficial. This would provide Atlas with more dynamic and varied driving scenarios, potentially reducing its overly conservative tendencies.

\textbf{Violation of Traffic Regulations. }Despite Atlas having learned to adhere to several traffic rules, it occasionally fails to comply with traffic light signals, as shown in Figure~\ref{fig:Violation}. Specifically, Atlas may proceed through intersections during a red light. This error stems from the model's lack of explicit traffic light information in its current framework. To mitigate this issue, integrating enhanced traffic-related data queries could be crucial. By providing Atlas with more explicit and detailed traffic signal information, we can improve its compliance with traffic laws and overall decision-making accuracy.

These findings highlight critical areas for further research and development. Enhancing the dataset and incorporating explicit models of traffic elements such as lights and signs are promising avenues for improving Atlas' performance and reliability.

\section{Border Impact}
Our work focuses on designing a VLM-based solution for autonomous driving.
Training our model using a large language model is often computationally demanding, which may result in significant environmental impacts.
Also, researchers with constrained computational resources may find it challenging to follow our work.
Except for these, there are no typical border impacts for our work.

\end{document}